\documentclass[runningheads]{llncs}

\usepackage{eccv}

\usepackage{eccvabbrv}

\usepackage{graphicx}
\usepackage{booktabs}

\usepackage[accsupp]{axessibility}  
\usepackage{booktabs}
\usepackage[table]{xcolor}
\usepackage{amssymb}
\usepackage{amsmath}
\usepackage{tabularx}
\usepackage[numbers]{natbib}
\usepackage{todonotes}
\usepackage{marvosym}
\definecolor{lighttext}{HTML}{888888} 

\newcommand{\methodname}{FLARE}
\newcommand{\g}[1]{\textcolor{gray!80}{#1}}

\usepackage{hyperref}

\usepackage{orcidlink}

\begin{document}

\title{\methodname: Learning Future-Aware Latent Representations from Vision-Language Models for Autonomous Driving} 

\titlerunning{Abbreviated paper title}

\author{
Chengen Xie$^{1,2,3}$ \quad
Chonghao Sima$^{3}$ \quad
Tianyu Li$^{2,3}$  \quad
Bin Sun$^{4}$ \quad
Junjie Wu$^{4}$ \quad 
\\
Zhihui Hao$^{4}$ \quad 
Hongyang Li$^{3~\textrm{\Letter}}$
\\[2mm]
}


\institute{\textsuperscript{\rm 1}  Shanghai Jiaotong University\quad
    \textsuperscript{\rm 2}  Shanghai Innovation Institute\\
    \textsuperscript{\rm 3}  OpenDriveLab at The University of Hong Kong \quad
    \textsuperscript{\rm 4} Li Auto Inc. \\
    \texttt{chengen.xie@opendrivelab.com}}

\maketitle

\begin{abstract}
  While Vision-Language Models (VLMs) offer rich world knowledge for end-to-end autonomous driving, current approaches heavily rely on labor-intensive language annotations (e.g., VQA) to bridge perception and control. This paradigm suffers from a fundamental mismatch between discrete linguistic tokens and continuous driving trajectories, often leading to suboptimal control policies and inefficient utilization of pre-trained knowledge. To address these challenges, we propose \textbf{\methodname} (\textbf{F}uture-aware \textbf{LA}tent \textbf{RE}presentation), a novel framework that activates the visual-semantic capabilities of pre-trained VLMs without requiring language supervision. Instead of aligning with text, we introduce a self-supervised \textbf{future feature prediction} objective. This mechanism compels the model to anticipate scene dynamics and ego-motion directly in the latent space, enabling the learning of robust driving representations from large-scale unlabeled trajectory data. Furthermore, we integrate Group Relative Policy Optimization (GRPO) into the planning process to refine decision-making quality. Extensive experiments on the NAVSIM benchmark demonstrate that \methodname\ achieves state-of-the-art performance, validating the effectiveness of leveraging VLM knowledge via predictive self-supervision rather than explicit language generation.
\end{abstract}

\section{Introduction}
\label{sec:intro}
\begin{figure}[t]
  \centering
  \includegraphics[width=\linewidth]{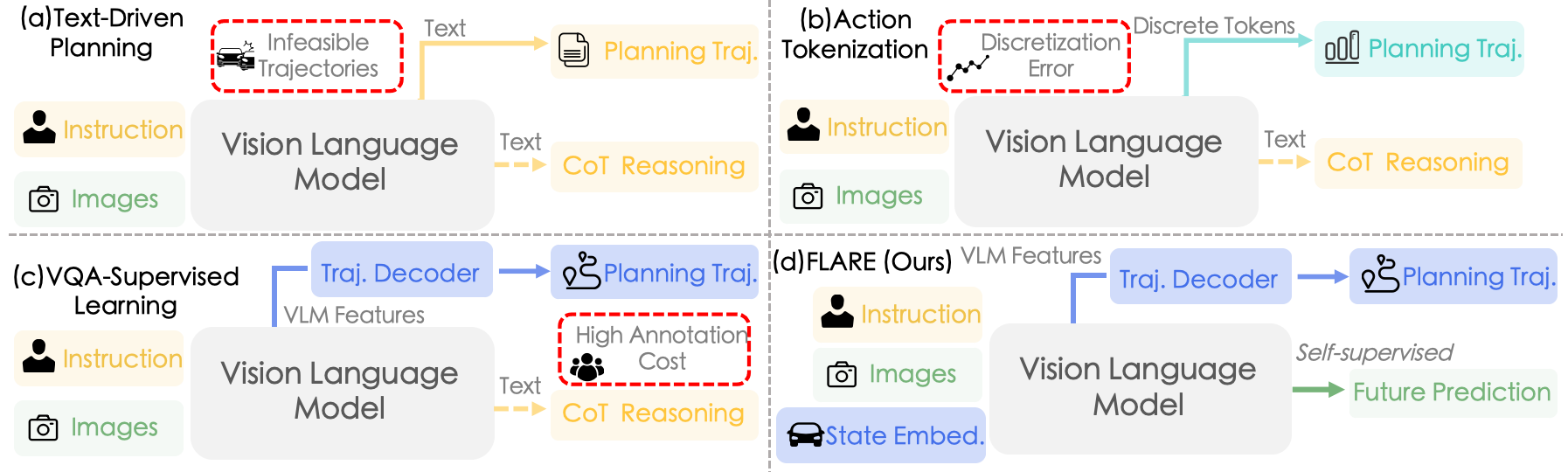} 
  \caption{\textbf{Comparison of VLM-based Autonomous Driving Paradigms.} 
  (a) Text-driven methods suffer from generating infeasible trajectories. 
  (b) Action tokenization introduces discretization errors and provides sparse supervision. 
  (c) Existing feature-based methods rely heavily on costly human annotations (VQA/Caption) to align representations. 
  (d) \textbf{\methodname{} (Ours)} introduces a self-supervised future prediction paradigm. By predicting future latent features instead of generating text, it achieves dense supervision and robust policy learning from unlabeled video data.}
  \label{fig:teaser}
\end{figure}

Recent end-to-end autonomous driving methods~\cite{uniad, jiang2023vad, hu2022stp3,sun2025sparsedrive}, which directly map raw sensor inputs to final trajectories, have demonstrated remarkable performance by learning human-like driving behaviors from large-scale datasets. Despite their success in common scenarios, these models face a fundamental limitation: the diversity of driving scenarios in training data remains substantially constrained compared to the complexity and variability inherent in real-world traffic conditions, inevitably imposing a performance ceiling when deployed in practice.
To address the challenge of handling rare long-tail scenarios, recent researches~\cite{ chen2024drivinggpt, hwang2024emma, jiang2025alphadrive, liu2025hybridvla, xu2025drivegpt4v2} has increasingly explored leveraging vision-language models (VLMs) pre-trained on large-scale internet data, which offer richer world knowledge and stronger generalization capabilities.
To better adapt VLMs to autonomous driving, numerous studies~\cite{chi2025impromptu,sima2024drivelm, tian2024drivevlm, zhou2025opendrivevlaendtoendautonomousdriving} have incorporated trajectory planning and driving reasoning tasks alongside visual question answering (VQA) during training, typically requiring extensive human annotations for scene descriptions, meta-actions, and driving rationales.

Despite competitive benchmark results, current VLM-based autonomous driving systems face critical hurdles regarding their training efficiency and control precision, as illustrated in Fig.~\ref{fig:teaser} (a-c). 
First, the annotation-intensive paradigm (Fig.~\ref{fig:teaser}c) yields diminishing returns. Relying on costly human annotations---such as VQA pairs and scene descriptions---often results in marginal performance gains~\cite{li2025recogdrive} while introducing semantic biases that misalign with actual driving behaviors.
Second, there is a fundamental mismatch between discrete linguistic reasoning and continuous motor control (Fig.~\ref{fig:teaser}a-b). Natural language is too coarse to capture tacit, non-verbalizable driving nuances, and the discrete tokenization inherent to VLMs is ill-suited for continuous action spaces. This leads to unstable trajectories and, critically, provides excessively sparse supervision signals that fail to fully leverage the massive parameter capacity of VLMs for fine-grained policy learning.

To systematically address these challenges, we propose \textbf{\methodname} (Fig.~\ref{fig:teaser}d), a novel framework that activates the rich world knowledge embedded in pre-trained VLMs without relying on explicit language annotations. 
First, to overcome the sparsity of action token supervision, we introduce a spatial future scene anticipation objective. Rather than aligning with discrete text, the model is trained to reconstruct fine-grained DINOv2 features of future frames. This self-supervised task forces the backbone to capture spatially-explicit scene dynamics and tacit driving knowledge, providing dense, continuous gradients that fully leverage the VLM's parameter capacity.
Second, to effectively utilize these representations, we design a two-stage fusion module that aligns visual features with the vehicle's ego-state embeddings, ensuring the planner is conditioned on both environmental context and physical status. Finally, we employ Group Relative Policy Optimization (GRPO) to fine-tune the trajectory head, optimizing the policy based on safety-critical rewards rather than simple imitation.

The main contributions of this work are:
\begin{enumerate}
\item  We propose a future feature prediction objective that reconstructs DINOv2 patches, enabling the VLM to learn robust motion representations from video data, mitigating the sparsity of pure action-token supervision.

\item We introduce a query-based fusion mechanism to align visual and state modalities, while employing GRPO to fine-tune the downstream trajectory policy. This design ensures the planned trajectories are both physically feasible and safety-optimized.

\item We achieve new state-of-the-art results on the NAVSIM~\cite{dauner2025navsim} benchmark among VLM-based methods under both SFT and RFT settings, while using only camera inputs.
\end{enumerate}

\section{Related Work}
\label{sec:related_work}

\subsection{End-to-End Autonomous Driving}
Conventional autonomous driving (AD) systems typically adopt modular pipelines, where perception, prediction, and planning components are optimized independently and integrated sequentially~\citep{chai2019multipath,li2024bevformer,liu2021multimodalmotionpredictionstacked}. To mitigate error propagation and enable joint optimization, recent approaches have shifted toward end-to-end learning that directly maps sensory inputs to planned trajectories. Transfuser~\cite{chitta2022transfuser} pioneered this direction by using a multi-task learning framework with shared feature extraction and task-specific heads. Subsequently, UniAD~\cite{uniad} and VAD~\cite{jiang2023vad} advanced this paradigm by constructing Bird's-Eye-View (BEV) representations from multi-camera inputs to perform perception, forecasting, and planning in a fully differentiable manner. To address the mode collapse often observed in regression-based planners, recent works such as VADv2~\cite{chen2024vadv2} and Hydra-MDP~\cite{li2024hydramdp} score predefined anchor trajectories to approximate multi-modal planning distributions. Similarly, DiffusionDrive~\cite{liao2024diffusiondrive} employs a truncated diffusion policy initialized from multi-mode anchors to efficiently capture complex action distributions with minimal denoising steps.
Despite these advances, pure end-to-end models remain constrained by training data coverage~\cite{tian2025simscalelearningdriverealworld}. When encountering long-tail scenarios outside the training distribution, they often exhibit performance degradation due to limited semantic generalization. This limitation motivates the integration of world knowledge from large-scale Vision-Language Models(VLMs) to enhance robustness in open-world driving environments.

\subsection{Vision-Language-Action Models for Driving}
Aligning semantic reasoning with physical action generation remains a fundamental challenge in integration Vision-Language Models (VLMs) into autonomous driving. Early research primarily utilized VLMs for scnene interpretation, such as captioning or question answering (e.g., DriveGPT4~\cite{xu2024drivegpt4}), or employed them as high-level planners to generate meta-actions for modular systems. However, these modular approaches are often constrained by non-differentiable interfaces, which prevent effective gradient back propagation and limit holistic optimization. 
More recently, unified Vision-Language-Action (VLA) models have emerged, directly mapping multimodal sensory inputs to driving trajectories. DriveMoE~\cite{yang2025drivemoe} adopts a hierarchical Mixture-of-Experts architecture to efficiently decouple planning from general reasoning, while AutoVLA’s~\cite{zhou2025autovla} integrates control into the language space via autoregressive action primitive tokenization. Similarly, ReCogDrive~\cite{li2025recogdrive} synergizes Chain-of-Thought reasoning with a diffusion-based trajectory generator to enhance interpretability. Most recently, DriveVLA-W0~\cite{li2025drivevlaw0} addresses the supervision bottleneck by employing a generative world model, training a diffusion transformer to predict future VAE latents.
Distinct from these generative approaches, our method bypasses pixel-level reconstruction by predicting DINOv2 feature patches, enabling the model to learn robust motion representations directly from video data. To bridge the gap between semantic reasoning and control, we employ a query-based fusion mechanism for modality alignment and further refine the policy using Group Relative Policy Optimization (GRPO), ensuring that the planned trajectories are both physically feasible and safety-optimized.

\section{Methodology}
\label{sec:method}

In this section, we introduce \methodname, an end-to-end autonomous driving framework designed to bridge the gap between the high-level semantic reasoning of Vision-Language Models (VLMs) and the low-level precision of continuous control. 
We begin by formalizing the trajectory generation task in Sec.~\ref{subsec:problem}. 
Sec.~\ref{subsec:overview} provides a holistic overview of the system's information flow. 
In Sec.~\ref{subsec:architecture}, we detail the architectural components, focusing on how semantic action queries condition the diffusion planner. 
Finally, Sec.~\ref{sec:training} outlines our two-stage optimization strategy, which combines supervised pre-training with a novel Group Relative Policy Optimization (GRPO) for safety refinement.

\subsection{Problem Formulation}
\label{subsec:problem}
We aim to build a generalizable end-to-end autonomous driving policy that unifies semantic scene understanding with precise motion planning. Formally, at each time step $t$, the ego-vehicle receives a set of heterogeneous observable inputs: a front-view image $I \in \mathbb{R}^{H \times W \times 3}$, a natural language navigation instruction $L$, and the vehicle's kinematic state $S = \{v_t, a_t, \tau_{t-k:t}\}$, which includes current velocity, acceleration, and historical trajectory.

The objective of our system is to learn a policy $\pi$ that maps these multimodal inputs to two complementary targets. The primary target is a safe, multi-modal future trajectory $\tau = \{(x_i, y_i, \theta_i)\}_{i=1}^{H}$ over $H$ time steps. To enforce physical scene understanding, we introduce a secondary self-supervised target: predicting the dense semantic feature map of the target frame, denoted as $\hat{\mathbf{F}} \in \mathbb{R}^{N_p \times d_f}$, where $N_p$ is the number of spatial patches and $d_f$ is the feature dimension. Thus, the problem is formulated as learning the joint mapping:
\begin{equation}
    f: \{I, L, S\} \rightarrow \{\tau, \hat{\mathbf{F}}\}.
\end{equation}

\begin{figure*}[t]
  \centering
  \includegraphics[width=1\linewidth]{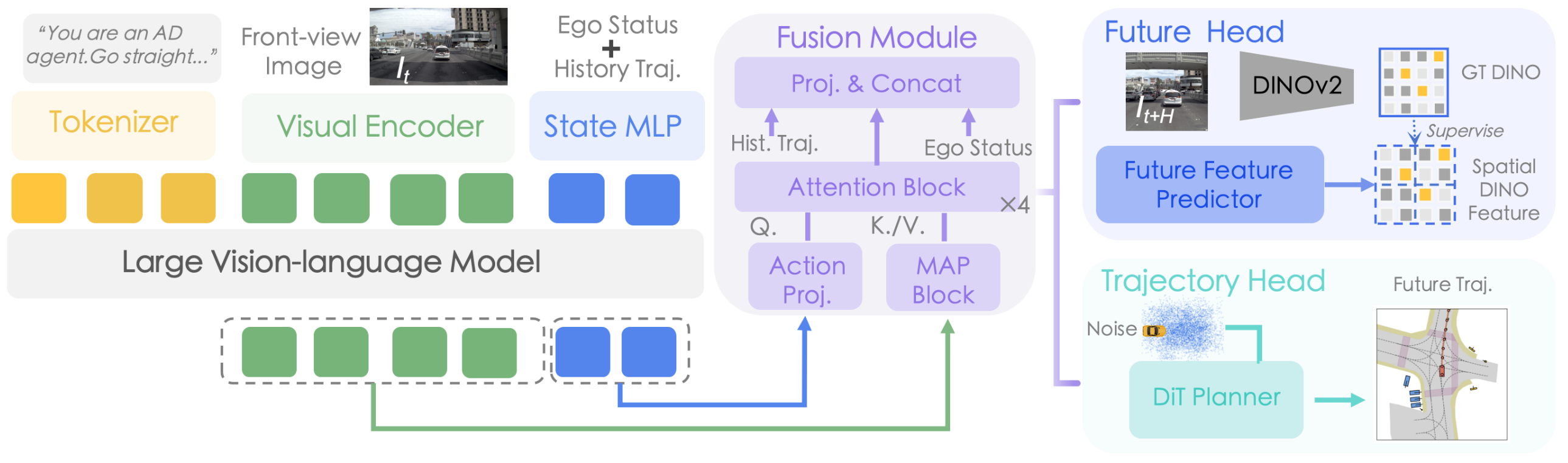}
  \caption{\textbf{The overall architecture of \methodname.} The framework processes heterogeneous inputs---front-view images, navigation commands, and ego-status---through a unified VLM backbone. A specialized Fusion Module aligns these multimodal features using query-based attention. The system employs a dual-head design: a \textbf{DiT Planner} for precise trajectory generation and a \textbf{Future Feature Predictor} that reconstructs DINOv2 patches to enforce physical understanding. Finally, the policy is refined via Group Relative Policy Optimization (GRPO) to ensure safety and feasibility.}
  \label{fig:architecture}
\end{figure*}

\subsection{Overview}
\label{subsec:overview}
As illustrated in Fig.~\ref{fig:architecture}, \methodname\ operates as a unified end-to-end policy network. The framework processes the formulated inputs through three distinct stages. First, a \textbf{Multimodal Semantic Perception} stage aligns visual, linguistic, and kinematic inputs into a unified token sequence, which is processed by a pre-trained Vision-Language Model (VLM) to extract rich semantic context. Second, a \textbf{Hierarchical Query-Based Fusion Module} compresses the high-dimensional VLM outputs into a compact, decision-centric latent space, filtering out task-irrelevant information. Finally, this latent representation drives a \textbf{Dual-Head Prediction} system: a Diffusion Transformer (DiT) planner that generates multi-modal future trajectories, and a Future Feature Predictor (FFP) that forecasts the semantic layout of the target frame to enforce physical scene understanding. The detailed architectural designs of these components are elaborated below.

\subsection{Architecture}
\label{subsec:architecture}

\paragraph{Multimodal Input Encoding.}
We align three distinct modalities into a unified token sequence fed to the VLM backbone. The visual encoder processes the front-view image $I$ into patch-level visual tokens, while the language tokenizer converts the navigation command $L$ into text tokens. Crucially, we encode the vehicle's dynamic context---comprising current velocity, acceleration, and historical trajectory---into a unified ego-state vector. This vector is projected by a State MLP with LayerNorm and GELU activations into $S$ learnable state tokens of dimension $d_{\text{llm}}$, which are appended to the visual-text token sequence before being passed to the VLM:
\begin{equation}
    \mathbf{H} = \mathrm{VLM}\bigl([T_{\text{vis}};\, T_{\text{lang}};\, T_{\text{state}}]\bigr) \in \mathbb{R}^{(K+M+S) \times d_{\text{llm}}},
\end{equation}
where $K$, $M$, and $S$ denote the number of visual, language, and state tokens respectively. To adapt the VLM to the driving domain while preserving its pre-trained reasoning capabilities, we apply Low-Rank Adaptation (LoRA) to the query, key, value, and output projection layers of the language model.

\paragraph{Hierarchical Query-Based Fusion Module.}
Directly decoding actions from the full high-dimensional VLM output $\mathbf{H}$ is both inefficient and prone to capturing task-irrelevant information. We therefore introduce a two-stage Multimodal Attention Pooling (MAP) mechanism to progressively compress $\mathbf{H}$ into a compact decision-centric representation.

In the first stage, we extract the visual tokens $\mathbf{H}_{\text{vis}}$ from the VLM output using positional masks. A visual MAP block then compresses these tokens into $N_v$ visual latents:
\begin{equation}
    \mathbf{V} = \mathrm{MAP}_{\text{vis}}(\mathbf{H}) \in \mathbb{R}^{N_v \times d},
\end{equation}
where $d$ denotes the hidden dimension of the fusion module. This step distills the scene-level semantic content from the full token sequence.

In the second stage, we extract the corresponding output state token $\mathbf{h}_{\text{state}}$---which encapsulates the driving command and ego-status---and project it via an Action Projector (a two-layer MLP with LayerNorm and GELU) to produce a state embedding:
\begin{equation}
    \mathbf{e}_{\text{state}} = \mathrm{ActionProj}(\mathbf{h}_{\text{state}}) \in \mathbb{R}^{d}.
\end{equation}
Subsequently, an action MAP block condenses the visual latents $\mathbf{V}$ into a single action decision vector $\mathbf{z}$. Crucially, this pooling process utilizes the state embedding $\mathbf{e}_{\text{state}}$ as the initialization query, effectively conditioning the visual aggregation on the vehicle's dynamic state and navigational intent:
\begin{equation}
    \mathbf{z} = \mathrm{MAP}_{\text{act}}(\mathbf{V} \mid \mathbf{e}_{\text{state}}) \in \mathbb{R}^{d}.
\end{equation}
The resulting fused vector $\mathbf{z}$ serves as the primary conditioning signal for the trajectory planner.

\paragraph{Trajectory Head (DiT Planner).}
To bridge the gap between high-level semantic reasoning and low-level control, we employ a conditional Diffusion Transformer (DiT)~\cite{peebles2023scalable} as the trajectory planner. 
Crucially, the planner is conditioned on the fused action query $\mathbf{z} = \mathrm{MAP}_{\text{act}}(\mathbf{V} \mid \mathbf{e}_{\text{state}})$, which encapsulates a hierarchical understanding of the driving scene. 
To further enforce kinematic feasibility, we explicitly re-inject the historical trajectory and current ego-status into the diffusion process. 
This design allows the DiT to denoise a sequence of future waypoints from a Gaussian initialization $\mathbf{x}_T$ to the precise trajectory $\tau$, effectively grounding the VLM's semantic intent in physical constraints. 
The model is trained via a standard diffusion objective:
\begin{equation}
    \mathcal{L}_{\text{traj}} = \mathbb{E}_{t,\mathbf{x}_0,\epsilon} \left[\|\epsilon - \epsilon_\theta(\mathbf{x}_t, t, \mathbf{z}, \mathbf{h}_{\text{traj}}, \mathbf{e}_{\text{state}})\|_2^2\right],
\end{equation}
where $\mathbf{x}_t$ denotes the noisy trajectory at step $t$, and the conditioning signals guide the generation toward safe and context-aware behaviors.

\paragraph{Future Feature Predictor.}
To ground the VLM's representations in physical dynamics, we introduce an auxiliary objective that predicts the high-level semantic features (DINOv2~\cite{oquab2023dinov2}) of the future frame. 
Specifically, we employ a set of learnable spatial queries $\mathbf{Q}_s$ that are first modulated by the action intent $\mathbf{z}$ to incorporate the planned maneuver. 
These intent-aware queries then retrieve relevant spatial context from the visual representations $\mathbf{V}$ via cross-attention to reconstruct the future scene state. 
Unlike pixel-level video generation, predicting semantic patches compels the model to internalize object permanence and motion logic while remaining invariant to nuisance factors like lighting changes. 
Crucially, conditioning on $\mathbf{z}$ ensures the predictor forecasts an action-conditional future, effectively serving as a "mental simulation" where the model anticipates how its specific decision influences the environment.

\subsection{Training Objectives}
\label{sec:training}

Our training pipeline is structured in two stages: initial multi-task supervised finetuning to establish a foundational driving policy, followed by Group Relative Policy Optimization (GRPO) to refine the policy for safety and comfort using non-differentiable metrics.

\paragraph{Stage 1: Joint Supervised Learning.}
We first optimize the model using a composite objective that balances trajectory generation with future scene prediction:
\begin{equation}
    \mathcal{L}_{\text{total}} = \mathcal{L}_{\text{traj}} + \lambda \mathcal{L}_{\text{future}},
\end{equation}
where $\lambda$ is a balancing coefficient.

The DiT planner is trained via standard denoising score matching. We sample a ground-truth trajectory $\tau_0$, a timestep $t \sim \mathcal{U}(0, T)$, and noise $\epsilon \sim \mathcal{N}(\mathbf{0}, \mathbf{I})$ to construct the noisy state $\mathbf{x}_t$. The model learns to recover the noise conditioned on the fused intent $\mathbf{z}$ and physical constraints. Trajectory Loss $\mathcal{L}_{\text{traj}}$ is formulated as:
\begin{equation}
    \mathcal{L}_{\text{traj}} = \mathbb{E}_{t,\tau_0,\epsilon} \left[\|\epsilon - \epsilon_\theta(\mathbf{x}_t, t, \mathbf{z}, \mathbf{h}_{\text{traj}}, \mathbf{e}_{\text{state}})\|_2^2\right].
\end{equation}

For the future feature predictor, we enforce alignment between the predicted features $\hat{\mathbf{F}}$ and the ground-truth DINOv2 features $\mathbf{F}_{\text{gt}}$ extracted from the future frame. The future prediction loss $\mathcal{L}_{\text{future}}$ combines magnitude regression and directional alignment:
\begin{equation}
    \mathcal{L}_{\text{future}} = \|\hat{\mathbf{F}} - \mathbf{F}_{\text{gt}}\|_1 + \alpha \left(1 - \frac{1}{N_p}\sum_{j=1}^{N_p} \mathrm{CosSim}(\hat{\mathbf{F}}_j, \mathbf{F}_{\text{gt},j})\right),
\end{equation}
where the cosine similarity term ensures the model captures the semantic layout of the scene (e.g., road boundaries, obstacles) regardless of absolute feature magnitude.

\paragraph{Stage 2: Policy Refinement via GRPO.}
Imitation learning is fundamentally bottlenecked by the quality of expert demonstrations and struggles to explicitly penalize rare but critical safety violations. To address this, we fine-tune the diffusion planner using Group Relative Policy Optimization (GRPO)~\cite{shao2024deepseekmath}, which optimizes a non-differentiable reward function without requiring a separate value network. During training, we evaluate each generated trajectory $\tau$ using a composite reward based on the PDM-Score~\cite{dauner2025navsim}: $R(\tau) = w_p R_{\text{progress}} + w_s R_{\text{safety}} + w_c R_{\text{comfort}}$. Here, $R_{\text{safety}}$ heavily penalizes collisions and near-misses (TTC), while $R_{\text{comfort}}$ constrains jerk and lateral acceleration to ensure human-like smoothness.

To optimize the diffusion policy, we approximate the intractable trajectory log-probability $\log \pi_\theta(\tau | \mathbf{z})$ by summing the log-probabilities of the Gaussian transitions across the reverse denoising chain. For each driving scenario, we sample a group of $G$ trajectories $\{\tau_i\}_{i=1}^G$ and compute their group-relative advantages $A_i = (R(\tau_i) - \bar{R}) / (\sigma_R + \varepsilon)$. Furthermore, to prevent policy collapse and maintain driving stability, we introduce a Behavior Cloning (BC) regularization loss $\mathcal{L}_{\text{BC}}$ in place of the standard KL divergence penalty. This auxiliary loss explicitly anchors the current policy to the frozen reference policy $\pi_{\text{ref}}$ from Stage 1. The final refinement objective maximizes the expected advantage subject to a PPO-style clipping constraint, augmented by the BC regularization:
\begin{equation}
    \mathcal{L}_{\text{total}} = -\frac{1}{G} \sum_{i=1}^{G} \min \left( r_i(\theta) A_i, \; \mathrm{clip}(r_i(\theta), 1-\epsilon, 1+\epsilon) A_i \right) + \lambda \mathcal{L}_{\text{BC}},
\end{equation}
where $r_i(\theta) = \pi_\theta(\tau_i | \mathbf{z}) / \pi_{\text{ref}}(\tau_i | \mathbf{z})$ is the likelihood ratio, and $\lambda$ controls the strength of the BC penalty.

\section{Experiments}
\label{sec:experiment}

\subsection{Experimental Setup}
\paragraph{Implementation Details.}
We adopt Qwen3-VL (4B variant) as our foundation visual-language model(VLM), comprising 4.6B parameters. In the SFT stage, We apply LoRA~\cite{hu2022lora} to the attention projection layers of the VLM backbone for parameter-efficient fine-tuning, with rank $r=8$ and scaling factor $\alpha=16$. We use separate learning rates for different parameter groups: $1 \times 10^{-4}$ for the fusion module and both prediction heads, and $3 \times 10^{-5}$ for the LoRA parameters. All learning rates follow a cosine decay schedule with a linear warmup over the first 5\% of training steps. The future prediction loss weight is set to $\lambda = 0.1$. Training is conducted for 80 epochs on 8 NVIDIA H100 GPUs with a per-GPU batch size of 8, 
yielding an effective batch size of 64. In the RFT stage, the VLM backbone is frozen; only the fusion module and DiT planner are updated. We set the learning rate to $2 \times 10^{-5}$ with the same cosine decay schedule. For each scene, we sample $G=16$ candidate trajectories using DDIM with $T'=5$ denoising steps, and evaluate each 
with the PDM-Score reward. The behavior cloning regularization weight is set to 0.1 to prevent policy collapse. The reinforcement fine-tuning runs for 15 epochs on the same hardware configuration.

\paragraph{Dataset.}
We train and evaluate our proposed method on the NAVSIM~\cite{dauner2025navsim, cao2025navsimv2} benchmark. NAVSIM is a planning-oriented autonomous driving dataset built upon OpenScene~\cite{openscene2023}, which itself is a redistribution of nuPlan~\cite{caesar2021nuplan}. The dataset is divided into two splits: \textit{navtrain}, containing 103,288 training frames, and \textit{navtest}, consisting of 12,146 evaluation frames. 

\paragraph{Evaluation Metrics.}
NAVSIM provides a non-reactive simulator that provides simulation-based metrics. The \textbf{PDM Score (PDMS)}, used in NAVSIM v1~\cite{dauner2025navsim}, aggregates five sub-metrics: No At-Fault Collisions (NC), Drivable Area Compliance (DAC), Time-to-Collision (TTC), Comfort (Comf.), and Ego Progress (EP). The \textbf{Extended PDM Score (EPDMS)}, 
introduced in NAVSIM v2~\cite{cao2025navsimv2}, further incorporates Traffic Light Compliance (TL), Lane Keeping (LK), and Extended Comfort (EC), offering a more comprehensive assessment of planning quality.

\begin{table*}[htbp]
    \centering
    \small
    \caption{\textbf{Performance comparison on NAVSIM v1 \textit{navtest}.} RFT results are reported for a comprehensive comparison. Symbols $^{\star}$ and $^{\dagger}$ denote fine-tuning on navtrain trajectory dataset and pre-training on external driving datasets, respectively.$^{\ddagger}$ denote using the best-of-N(N=6) strategy. Bold indicates the best performance under SFT and RFT settings.}
    \vspace{-5pt}
    \setlength{\tabcolsep}{3pt}
    \begin{tabularx}{\textwidth}{@{}X|c|cc|ccc|c@{}}
        \toprule
        Method &  Sensors & NC$\uparrow$ & DAC$\uparrow$ & TTC$\uparrow$ & Comf. $\uparrow$ & EP$\uparrow$ & PDMS$\uparrow$ \\
        \midrule
        Constant Velocity & - & 68.0 & 57.8 & 50.0 & 100.0 & 19.4 & \cellcolor{gray!30} 20.6 \\
        Ego Status MLP & - & 93.0 & 77.3 & 83.6 & 100.0 & 62.8 & \cellcolor{gray!30} 65.6 \\        
        \midrule
        Hydra-MDP-$\mathcal{V}_{\text{8192}}$~\citep{li2024hydramdp} & 3$\times$Cam$+$L &   97.9 & 91.7 & 92.9 & 100.0 & 77.6 & \cellcolor{gray!30} 83.0 \\
        UniAD~\citep{uniad} & 6$\times$Cam    & 97.8 & 91.9 & 92.9 & 100.0 & 78.8 & \cellcolor{gray!30} 83.4 \\
        LTF~\citep{chitta2022transfuser} & 3$\times$Cam    & 97.4 & 92.8 & 92.4 & 100.0 & 79.0 & \cellcolor{gray!30} 83.8 \\
        TransFuser~\citep{chitta2022transfuser} & 3$\times$Cam$+$L   & 97.7 & 92.8 & 92.8 & 100.0 & 79.2 & \cellcolor{gray!30} 84.0 \\
        PARA-Drive~\citep{weng2024paradrive} & 6$\times$Cam  & 97.9 & 92.4 & 93.0 & 99.8 & 79.3 & \cellcolor{gray!30} 84.0 \\
        Epona~\citep{zhang2025epona} & 3$\times$Cam    & 97.9 & 95.1 & 93.8 & 99.9 & 80.4 & \cellcolor{gray!30} 86.2 \\
        Hydra-MDP-$\mathcal{V}_{\text{8192}}$-W-EP~\citep{li2024hydramdp} & 3$\times$Cam$+$L  & 98.3 & 96.0 & 94.6 & 100.0 & 78.7 & \cellcolor{gray!30} 86.5 \\
        ARTEMIS~\citep{feng2025artemis} & 3$\times$Cam$+$L & 98.3 & 95.1 & 94.3 & 100.0 & 81.4 & \cellcolor{gray!30} 87.0 \\
        DiffusionDrive~\citep{liao2024diffusiondrive} & 3$\times$Cam$+$L & 98.2 & 96.2 & 94.7 & 100.0 & 82.2 & \cellcolor{gray!30} 88.1 \\
        WoTE~\citep{li2025wote} & 3$\times$Cam$+$L & 98.5 & 96.8 & 94.9 & 99.9 & 81.9 & \cellcolor{gray!30} 88.3 \\
        \midrule
        \multicolumn{8}{@{}l}{\raggedright \textbf{VLMs-based Methods~(SFT)}} \\
        \g{AutoVLA~3B$^{\dagger}$\citep{zhou2025autovla}} & \g{3$\times$Cam}   &  \g{96.9} & \g{92.4} & \g{88.1} & \g{99.1} & \g{75.8} & \g{80.5} \\
        QwenVL3-4B$^{\star}$~\citep{Qwen2.5-VL} & 1$\times$Cam &  96.0 & 91.0 & 90.8 & 99.7 & 76.3 & \cellcolor{gray!30} 80.8 \\
        ReCogDrive-2B~\citep{li2025recogdrive} & 3$\times$Cam  & 97.6 & 93.1 & 92.7 & 100.0 & 79.1 & \cellcolor{gray!30} 84.1 \\
        \g{ReCogDrive-2B$^{\dagger}$~\citep{li2025recogdrive}} & \g{3$\times$Cam}  &  \g{98.1} & \g{94.7} & \g{94.2} & \g{100.0} & \g{80.9} & \g{86.5} \\
        \g{ReCogDrive-8B$^{\dagger}$~\citep{li2025recogdrive}} & \g{3$\times$Cam}  &  \g{98.3} & \g{95.1} & \g{94.3} & \g{100.0} & \g{81.1} & \g{86.8} \\
        \midrule
        \rowcolor{gray!30}\methodname-4B~(ours) & 1$\times$Cam  & \textbf{98.2} & \textbf{95.0} &\textbf{ 93.9} & \textbf{100.0} & \textbf{81.8} & \cellcolor{gray!30} \textbf{86.9} \\
        \midrule
        \multicolumn{8}{@{}l}{\raggedright \textbf{VLMs-based Methods~(RFT)}} \\
        AutoVLA-3B~\citep{zhou2025autovla} & 3$\times$Cam  &  98.4 & 95.6 & \textbf{98.0} & 99.9 & 81.9 & \cellcolor{gray!30} 89.1 \\
        DriveVLA-W0~\citep{li2025drivevlaw0} & 1$\times$Cam  &  \textbf{98.7} & \textbf{99.1} & 95.3 & 99.3 & 83.3 & \cellcolor{gray!30} 90.2 \\
        ReCogDrive-8B~\citep{li2025recogdrive} & 3$\times$Cam  &  97.8 & 97.7 & 94.9 & \textbf{100.0} & 86.3 & \cellcolor{gray!30} 90.5 \\
        ReCogDrive-2B~\citep{li2025recogdrive} & 3$\times$Cam  &  97.9 & 97.3 & 94.9 & \textbf{100.0} & \textbf{87.3} & \cellcolor{gray!30} 90.8 \\
        \g{AutoVLA-3B$^{\ddagger}$~\citep{zhou2025autovla}} & \g{3$\times$Cam}   &  \g{99.1} & \g{97.1} & \g{97.1} & \g{100.0} & \g{87.6} & \g{92.1} \\
        \g{DriveVLA-W0$^{\ddagger}$~\citep{li2025drivevlaw0}} & \g{3$\times$Cam}  &  \g{99.3} &\g{97.4} & \g{97.0} & \g{99.9} & \g{88.3} & \g{93.0} \\
        \midrule
        \rowcolor{gray!30}\methodname-4B~(ours) & 1$\times$Cam  & 98.5 & 98.4 & 96.0 & \textbf{100.0} & 86.0 & \cellcolor{gray!30} \textbf{91.4} \\
        \bottomrule
    \end{tabularx}
    \label{tab:main_results_on_pdms_rl}
    \vspace{-5pt}
\end{table*}
\subsection{Main Results}
We present a comprehensive quantitative evaluation of our method on \textit{navtest} dataset with both NAVSIM v1 and v2 metrics, detailed in Table~\ref{tab:main_results_on_pdms_rl} and Table~\ref{tab:main_results_on_epdms}, respectively. We compare our approach against state-of-the-art vision-based and multi-modal driving models under both Supervised Fine-Tuning (SFT) and Reinforcement Fine-Tuning (RFT) settings.

Under the SFT setting on NAVSIM v1 (Table~\ref{tab:main_results_on_pdms_rl}), our \methodname-4B achieves a PDMS score of 86.9, slightly surpassing the best-performing ReCogDrive-8B (86.8) while using only half the model parameters (4B vs. 8B). Notably, our method attains this performance without relying on external pre-training datasets, unlike ReCogDrive and AutoVLA which leverage additional driving data. Compared to the baseline QwenVL3-4B trained on the same navtrain trajectory dataset, our approach yields a substantial improvement of +6.1 PDMS (86.9 vs. 80.8), demonstrating the effectiveness of our proposed framework. Furthermore, our vision-only method significantly narrows the gap with traditional end-to-end approaches such as DiffusionDrive (88.1 PDMS) and WoTE (88.3 PDMS) that utilize both image and LiDAR inputs, achieving competitive performance with a much simpler single-camera configuration.

Under the RFT setting, our \methodname-4B further improves to 91.4 PDMS, outperforming ReCogDrive-2B (90.8) and ReCogDrive-8B (90.5) by clear margins. While AutoVLA-3B and DriveVLA-W0 report higher scores of 92.1 and 93.0 PDMS respectively, these results rely on the best-of-N (N=6) sampling strategy, which introduces additional computational overhead at inference time. Under fair single-sample evaluation, our method achieves the best performance among all VLM-based approaches. Across all sub-metrics, our approach maintains a perfect Comfort score of 100 and achieves competitive performance on NC (98.5), DAC (98.4), and TTC (96.0), indicating robust and safe driving behavior.

\begin{table*}[ht]
  \centering
  \caption{\textbf{Performance comparison on NAVSIM v2 \textit{navtest} with extended metrics.} Our method achieves state-of-the-art EPDMS while maintaining competitive scores across all sub-metrics. Bold indicates the best performance.}
  \label{tab:main_results_on_epdms}
  \small
  \scalebox{1}{
  \begin{tabular}{l|cccc|ccccc|c}
    \toprule
    Method     & NC$\uparrow$ & DAC$\uparrow$ & EP$\uparrow$ & TTC$\uparrow$ & HC$\uparrow$ & TL$\uparrow$ & DDC$\uparrow$ & LK$\uparrow$ & EC$\uparrow$ & EPDMS$\uparrow$ \\
    \midrule
    Transfuser~\citep{chitta2022transfuser}    & 97.7 & 92.8 & 79.2 & 92.8 & \textbf{100.0}   & 99.9 & 98.3 & 67.6 & 95.3 & \cellcolor{gray!30}77.8 \\
    VADv2~\citep{chen2024vadv2}         & 97.3 & 91.7 & 77.6 & 92.7 & \textbf{100.0}   & 99.9 & 98.2 & 66.0 & 97.4 & \cellcolor{gray!30}76.6 \\
    Hydra-MDP~\citep{li2024hydramdp}     & 97.5 & 96.3 & 80.1 & 93.0 & \textbf{100.0}   & 99.9 & 98.3 & 65.5 & 97.4 & \cellcolor{gray!30}79.8 \\
    Hydra-MDP++~\citep{li2025hydramdp++}   & 97.9 & 96.5 & 79.2 & 93.4 & \textbf{100.0}   & \textbf{100.0} & 98.9 & 67.2 & 97.7 & \cellcolor{gray!30}80.6 \\
    ARTEMIS~\citep{feng2025artemis}           & 98.3 & 95.1 & 81.5 & 97.4 & \textbf{100.0} & 99.8 & 98.6 & 96.5 & \textbf{98.3} & \cellcolor{gray!30}83.1 \\
    ReCogDrive-8B~\citep{li2025recogdrive}  & 98.3 & 95.2 & 87.1 & 97.5 & 98.3 & 99.8 & \textbf{99.5} & 96.6 & 86.5 & \cellcolor{gray!30}83.6 \\
    DiffusionDrive~\citep{liao2024diffusiondrive}  & 98.0 & 96.0 & 87.7 & 97.1 & 98.3 & 99.8 & \textbf{99.5} & \textbf{97.2} & 87.6 & \cellcolor{gray!30}84.3 \\
    DriveVLA-W0~\citep{li2025drivevlaw0}  & 98.5 & \textbf{99.1} & 86.4 & 98.1 & 97.9 & 99.7 & 98.0 & 93.2 & 58.9 & \cellcolor{gray!30}86.1 \\
    \midrule
    \methodname-4B~(ours)  & \textbf{98.8} & 96.6& \textbf{87.9} & \textbf{98.2} & 98.4 & 99.9 & \textbf{99.5} & 95.8 & 87.5 & \cellcolor{gray!30}\textbf{86.3} \\
    \bottomrule
  \end{tabular}}
\end{table*}

To further validate the robustness of our approach, Table~\ref{tab:main_results_on_epdms} extends our evaluation to the NAVSIM v2 benchmark using the comprehensive EPDMS metric. Our \methodname-4B achieves state-of-the-art performance with an EPDMS of 86.3, outperforming strong baselines including DriveVLA-W0 (86.1) and DiffusionDrive (84.3). It is worth noting that while DriveVLA-W0 shows a slight edge in DAC (99.1), it suffers from a severe degradation in Extended Comfort (EC, 58.9). In contrast, our method maintains a highly balanced and safe driving profile, securing the highest scores in NC (98.8), EP (87.9), and TTC (98.2), alongside a strong EC score of 87.5. This demonstrates that our approach not only excels in fundamental navigation tasks but also strictly adheres to complex environmental constraints.

\subsection{Ablation Results}
We conduct comprehensive ablation studies to validate the effectiveness of our key design choices, including the future feature prediction task and the fusion module.

\begin{table}[htbp]
    \centering
    \small
    \caption{\textbf{Ablation study on Future Feature Prediction.} Comparing different prediction targets reveals that predicting spatial DINO features yields the highest PDMS by capturing fine-grained scene dynamics.}
    \vspace{-5pt}
    \setlength{\tabcolsep}{4pt}
    \begin{tabular}{l|l|cc|ccc|c}
        \toprule
        ID & Prediction Target & NC$\uparrow$ & DAC$\uparrow$ & TTC$\uparrow$ & Comf.$\uparrow$ & EP$\uparrow$ & PDMS$\uparrow$ \\
        \midrule
        1 & None (Pure Traj. Prediction) & 97.6 & 92.5 & 92.8 & \textbf{100.0} & 78.1 & \cellcolor{gray!30}83.4 \\
        2 & Image Pixels & 97.7 & 93.7 & 93.5 & \textbf{100.0} & 78.9 & \cellcolor{gray!30}84.7 \\
        3 & Global DINO Feature & 98.1 & 94.2 & 93.6 & \textbf{100.0} & 80.8 & \cellcolor{gray!30}85.9 \\
        \midrule
        \rowcolor{gray!15}
        4 & \textbf{Spatial DINO (Ours)} & \textbf{98.2} & \textbf{95.0} &\textbf{ 93.9} & \textbf{100.0} & \textbf{81.8} & \cellcolor{gray!30} \textbf{86.9} \\
        \bottomrule
    \end{tabular}
    \label{tab:ablation_prediction}
\end{table}
\paragraph{Effect of Future Feature Prediction.} As shown in Table~\ref{tab:ablation_prediction},  we investigate different prediction targets for the future feature prediction task. The baseline model without any future prediction (ID 1) achieves 83.4 PDMS, serving as a lower bound. Introducing image pixel prediction (ID 2) brings a modest improvement of +1.3 PDMS, indicating that explicit future reasoning benefits planning. Replacing pixel-level targets with global DINO features (ID 3) further boosts performance to 85.9 PDMS, as semantic features provide more structured supervision than raw pixels. Our proposed spatial DINO prediction (ID 4) achieves the best performance of 86.9 PDMS, outperforming the global feature variant by +1.0 PDMS. This demonstrates that preserving spatial granularity in future feature prediction enables the model to capture fine-grained scene dynamics, which is crucial for accurate trajectory planning.

\begin{table}[htbp]
    \centering
    \small
    \caption{\textbf{Ablation study on Fusion Module.} Evaluating different fusion strategies shows that our two-stage MAP fusion and ego-state integration are both indispensable for optimal planning performance.}
    \vspace{-5pt}
    \setlength{\tabcolsep}{4pt}
    \begin{tabular}{l|l|cc|ccc|c}
        \toprule
        ID & Fusion Variant & NC$\uparrow$ & DAC$\uparrow$ & TTC$\uparrow$ & Comf.$\uparrow$ & EP$\uparrow$ & PDMS$\uparrow$ \\
        \midrule
        1 & w/o MAP Fusion & 97.7 & 93.8 & 93.4 & \textbf{100.0} & 79.6 & \cellcolor{gray!30}85.0 \\
        2 & w/o State Injection & 97.6 & 93.8 & 93.0 & \textbf{100.0} & 79.5 & \cellcolor{gray!30}84.8 \\
        \midrule
        \rowcolor{gray!15}
        3 & \textbf{Two-Stage MAP (Ours)} & \textbf{98.2} & \textbf{95.0} &\textbf{ 93.9} & \textbf{100.0} & \textbf{81.8} & \cellcolor{gray!30} \textbf{86.9} \\
        \bottomrule
    \end{tabular}
    \label{tab:ablation_fusion}
\end{table}
\paragraph{Effect of Fusion Module.} Table~\ref{tab:ablation_fusion} examines the contribution of our two-stage MAP design within the fusion module. Replacing the MAP fusion with global average pooling (ID 1) leads to a performance drop of -1.9 PDMS (85.0 vs. 86.9), suggesting that spatially-aware feature aggregation is essential for effective multi-modal reasoning. Replacing the ego-state embedding with an all-zero vector (ID 2) results in a similar degradation to 84.8 PDMS, highlighting the importance of conditioning the planning process on vehicle dynamics. Our full model with the complete fusion module (ID 3) achieves the best results across all metrics, validating that both components are complementary and indispensable for optimal performance.

\begin{figure*}[t]
  \centering
  \includegraphics[width=1\linewidth]{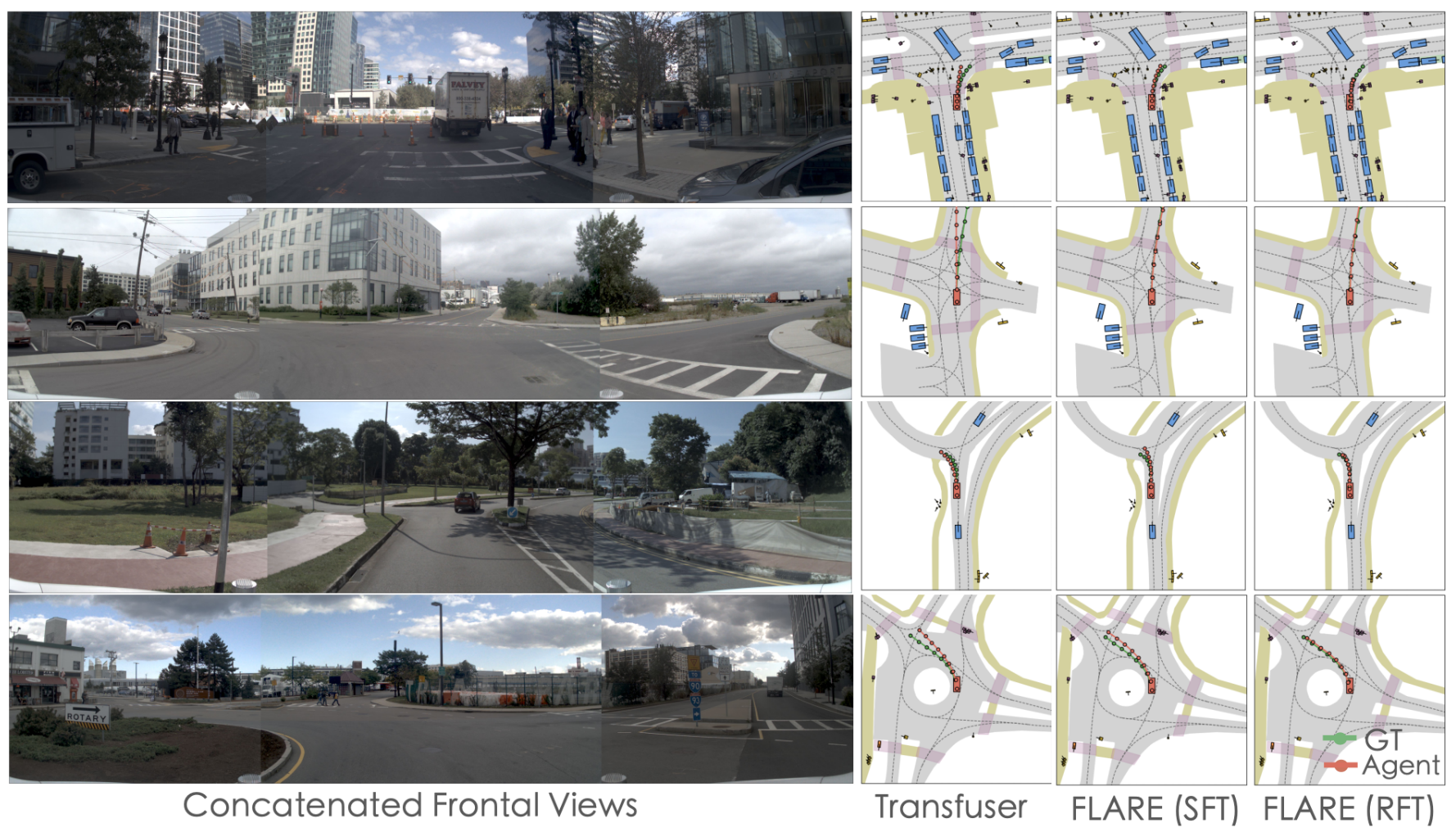}
  \caption{\textbf{Qualitative comparison on challenging \textit{navtest} scenarios.} Compared to baselines, our full method (\methodname-RFT) demonstrates superior safety awareness by proactively yielding in critical situations (e.g., Row 1) and generates smooth, kinematically feasible trajectories perfectly aligned with lane centers in complex topologies (e.g., Rows 3-4).}
  \label{fig:qualitative}
\end{figure*}

\subsection{Qualitative Analysis}
\label{subsec:qualitative}
To intuitively understand the impact of our proposed components, we visualize the planning results in diverse, challenging driving scenarios in Fig.~\ref{fig:qualitative}. We compare our full method (\methodname-RFT) against the baseline Transfuser and our intermediate Supervised Fine-Tuning stage (\methodname-SFT). It is worth noting that these scenarios are \textbf{not cherry-picked}; they represent common yet critical situations including urban intersections, curved roads, and roundabouts.

\paragraph{Safety-Critical Decision Making.}
The first row of Fig.~\ref{fig:qualitative} highlights the critical advantage of our GRPO-based refinement. In this scenario, the ego-vehicle encounters a congested intersection with a truck obstructing the path.
While the SFT model (middle column) blindly mimics the expert's tendency to maintain speed, resulting in a risky trajectory that could lead to a collision, our RFT model (right column) exhibits superior safety awareness.
Notably, the RFT agent predicts a \textbf{shorter trajectory length}, indicating a proactive deceleration and yielding behavior. This demonstrates that the safety reward in GRPO effectively suppresses the aggressive behaviors often found in imitation learning, teaching the agent to prioritize collision avoidance over mere route progress.

\paragraph{Complex Topology Handling.}
The subsequent rows demonstrate the model's robustness in complex topologies.
In the curved road scenario (Row 3) and the roundabout entry (Row 4), the Transfuser baseline struggles with lane centering, often drifting towards lane boundaries. The SFT model improves upon this but still exhibits minor jitter.
In contrast, \methodname-RFT generates smooth, kinematically feasible trajectories that align perfectly with the lane centerlines. This confirms that our Future Feature Prediction Head successfully helps the model capture the underlying scene geometry and dynamics, while the RL refinement ensures the generated plans are stable and comfortable.

\section{Conclusion}
\label{sec:conclusion}
In this paper, we introduced \methodname, a novel end-to-end autonomous driving framework that unlocks the visual-semantic reasoning capabilities of pre-trained Vision-Language Models (VLMs) without relying on labor-intensive language annotations. By formulating a self-supervised future spatial feature prediction objective, \methodname \ learns robust, continuous driving representations directly from unlabeled video data, effectively bridging the fundamental gap between discrete linguistic tokens and continuous motor control. Furthermore, our proposed two-stage MAP fusion mechanism seamlessly aligns visual context with ego-state dynamics, while the integration of Group Relative Policy Optimization (GRPO) refines the planning policy to strictly prioritize safety and comfort. Extensive evaluations on the NAVSIM benchmark demonstrate that \methodname\ achieves state-of-the-art performance among VLM-based methods in both SFT and RFT settings, utilizing only single-camera input. Ultimately, our work highlights a highly efficient paradigm: leveraging VLM world knowledge through predictive self-supervision, rather than explicit text generation, offers a highly effective and scalable pathway for autonomous driving.
\section{Limitations and Future Work}
\label{sec:limitation}
While \methodname\ demonstrates strong performance and training efficiency, several limitations remain to be addressed in future research. First, our current study does not explicitly explore the effects of data scaling. Given that our self-supervised future feature prediction objective is inherently designed to leverage unlabeled video data, investigating how \methodname{}'s performance scales with massive, diverse driving datasets is a critical next step. Second, the generalization capabilities of our framework have primarily been validated on the NAVSIM benchmark. Its robustness and adaptability have yet to be extensively tested across a broader range of autonomous driving benchmarks or in highly interactive closed-loop simulations. Future work will focus on scaling up the SFT training data corpus and deploying FLARE across multiple diverse environments to fully assess its open-world generalization and zero-shot capabilities.
\section*{Acknowledgments}
This work is in part supported by the JC STEM Lab of Autonomous Intelligent Systems funded by The Hong Kong Jockey Club Charities Trust. We also appreciate the generous research sponsor from Li Auto.

We extend our gratitude to all the members from OpenDriveLab and Li Auto for their profound support.

%
%
\bibliographystyle{splncs04}
\bibliography{main}
\end{document}